\documentclass[10pt,twocolumn,letterpaper]{article}

\usepackage[pagenumbers]{cvpr} % To force page numbers, e.g. for an arXiv version

\definecolor{cvprblue}{rgb}{0.21,0.49,0.74}
\usepackage[pagebackref,breaklinks,colorlinks,allcolors=cvprblue]{hyperref}
\usepackage{lib}

\usepackage{algorithm}
\usepackage{algorithmic}
\usepackage{lipsum} 
\usepackage{amsmath}
\usepackage{hyperref}
\usepackage{booktabs}
\usepackage{xcolor}
\usepackage{colortbl} % For cell colors

\usepackage{pifont}

\newcommand{\ceil}[1]{\left\lceil #1 \right\rceil}

\newcommand{\ours}{DSD}

\definecolor{colorfid}{rgb}{0.2, 0.9, 0.85} % RGB values between 0 and 1
\definecolor{colorclip}{rgb}{0.5, 0.9, 0.2}
\definecolor{colorlpips}{rgb}{1.0, 0.4, 0.1}
\newcommand{\colorCLIP}[1]{\cellcolor{colorclip!#1}} % Green for CLIP-SIM (higher is better)
\title{Diverse Score Distillation}

\author{Yanbo Xu $^{1}$ \hspace{12pt}
Jayanth Srinivasa $^{2}$ \hspace{12pt}
Gaowen Liu $^{2}$ \hspace{12pt}
Shubham Tulsiani $^{1}$
\\
$^{1}$ Carnegie Mellon University \hspace{12pt}
$^{2}$ Cisco Research 
\\
\textit{Project Page}: \href{https://billyxyb.github.io/Diverse-Score-Distillation/}{billyxyb.github.io/Diverse-Score-Distillation}\\
}

\begin{document}
% \maketitle
% from https://tex.stackexchange.com/questions/55764/input-a-figure-between-title-and-body-in-twocolumn-form
\twocolumn[{%
\renewcommand\twocolumn[1][]{#1}%
\maketitle
\begin{center}
    \centering
    \captionsetup{type=figure}
    \includegraphics[width=1.0\textwidth,height=8cm]{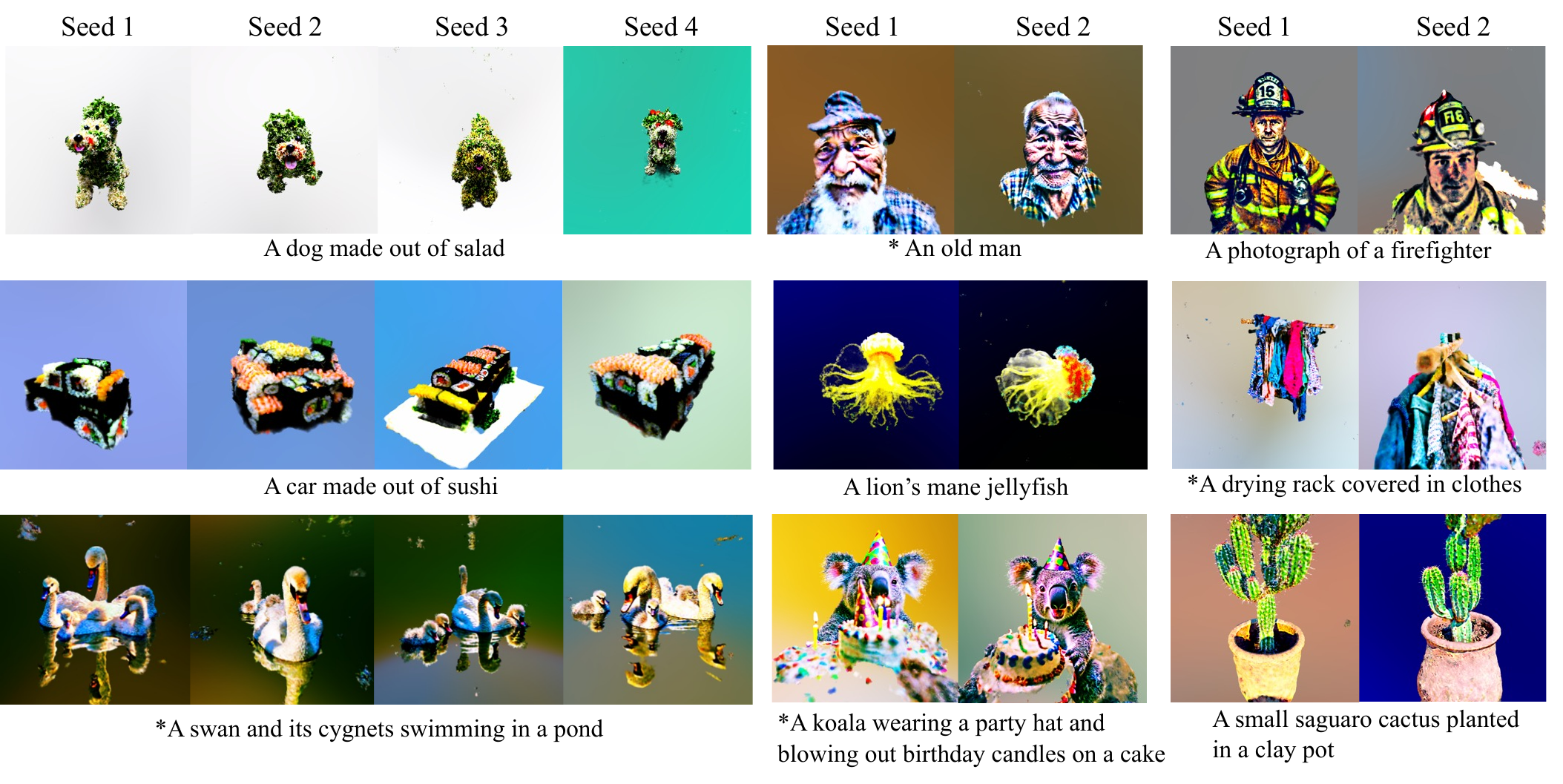}
    \captionof{figure}{ \textbf{Diverse Score Distillation.} We present a sampling-inspired score distillation formulation that allows obtaining diverse (3D) outputs via different initial optimization seeds. \small{* "A DSLR photo of"}.  
    }
\end{center}%
}]

\begin{abstract}
Score distillation of 2D diffusion models has proven to be a powerful mechanism to guide 3D optimization, for example enabling text-based 3D generation or single-view reconstruction.  A common limitation of existing score distillation formulations, however, is that the outputs of the (mode-seeking) optimization are limited in diversity despite the underlying diffusion model being capable of generating diverse samples. In this work, inspired by the sampling process in denoising diffusion, we propose a score formulation that guides the optimization to follow generation paths defined by random initial seeds, thus ensuring diversity. We then present an approximation to adopt this formulation for scenarios where the optimization may not precisely follow the generation paths (\eg a 3D representation whose renderings evolve in a co-dependent manner). We showcase the applications of our `Diverse Score Distillation' (DSD) formulation across tasks such as 2D optimization, text-based 3D inference, and single-view reconstruction. We also empirically validate DSD against prior score distillation formulations and show that it significantly improves sample diversity while preserving fidelity.

\end{abstract}

\section{Introduction}
\label{sec:intro}

The impressive progress in generative AI has helped democratize the ability to create visual content. In particular, recent image~\cite{stablediffusion} or video~\cite{videodiffusion} generation models allow end-users to easily create  (diverse and photorealistic) visual output from just a text prompt, for example enabling one to synthesize images of an avocado chair or a teddybear skating in Times Square. This success of 2D generative models, however, has not yet been matched by their 3D counterparts, and the goal of inferring diverse and high-fidelity 3D outputs from just a text prompt or an image remains an elusive goal. One key bottleneck towards this is the availability of 3D data. Although we have witnessed promising advances from generative 3D~\cite{ntavelis2023autodecoding, chen2024v3d} or multi-view~\cite{shi2023MVDream, kant2024spad} approaches that leverage synthetic 3D~\cite{deitke2023objaverse} or real-world multi-view datasets~\cite{reizenstein2021common}, their diversity and scale is still short of the 2D datasets that empower complex and photorealistic generation.

As an alternate paradigm for 3D generation without learning generative 3D (or multi-view) models, several approaches have explored mechanisms to `distill' pre-trained large-scale 2D generative models for 3D inference. 
Following DreamFusion~\cite{poole2022dreamfusion}, which introduced a `score distillation sampling' (SDS) formulation to approximate log-likelihood gradients from a diffusion model, these approaches cast 3D inference as a (2D diffusion-guided) optimization task and leverage 2D diffusion models to obtain gradients for renderings of 3D representations being optimized. While follow-up methods in this paradigm have since improved various aspects of this pipeline, these methods are all fundamentally \emph{`mode seeking'}, and thus (unlike diffusion-based generation) exhibit limited diversity in their inferred 3D representations (see 
\cref{fig:3D compare}). 

In this work, we present an alternate formulation for distilling diffusion models that overcomes this limitation, and allows diverse (3D) generation. Our approach is inspired by the ODE perspective on sampling from diffusion models~\cite{song2020score} which highlights that a trained model can be viewed as inducing a learned ODE that maps noise samples to data, and different starting points for the ODEs (\ie different initial noise samples) yield diverse data samples. Building on this insight, we derive a gradient that allows an optimization to `follow' an ODE, and enables diverse optimization outputs by simply specifying different ODE (noise) initializations. We first validate our approach for 2D image generation via optimization and show that, akin to DDIM sampling, it allows diverse and high-fidelity generation.

However, when naively applying this ODE-based formulation for 3D optimization, we find that it does not result in plausible 3D outputs. We show that this is because while a 2D optimization process can perfectly follow a specified (2D) ODE, the 3D optimization process seeks to infer a 3D representation that follows such ODEs across multiple views, and inherently cannot do so perfectly. We then generalize our diffusion distillation formulation to allow for such drift, and show that it allows high-fidelity and diversity for 3D (and 2D) inference. To better highlight the connections (and key differences) of our formulation with prior score distillation objectives, we also present a unifying perspective across existing methods.

We validate our approach for text-based-3D optimization using a pre-trained 2D diffusion model and show that our formulation yields similar (or better) generations in terms of quality compared to the state-of-the-art, while allowing significantly more diverse outputs. In addition, we also showcase our approach's ability to guide single-view 3D inference via distilling novel-view diffusion priors, and in particular, find that it allows multi-modal 3D outputs. 

\begin{figure}[]
  \centering
   \includegraphics[width=1.0\linewidth]{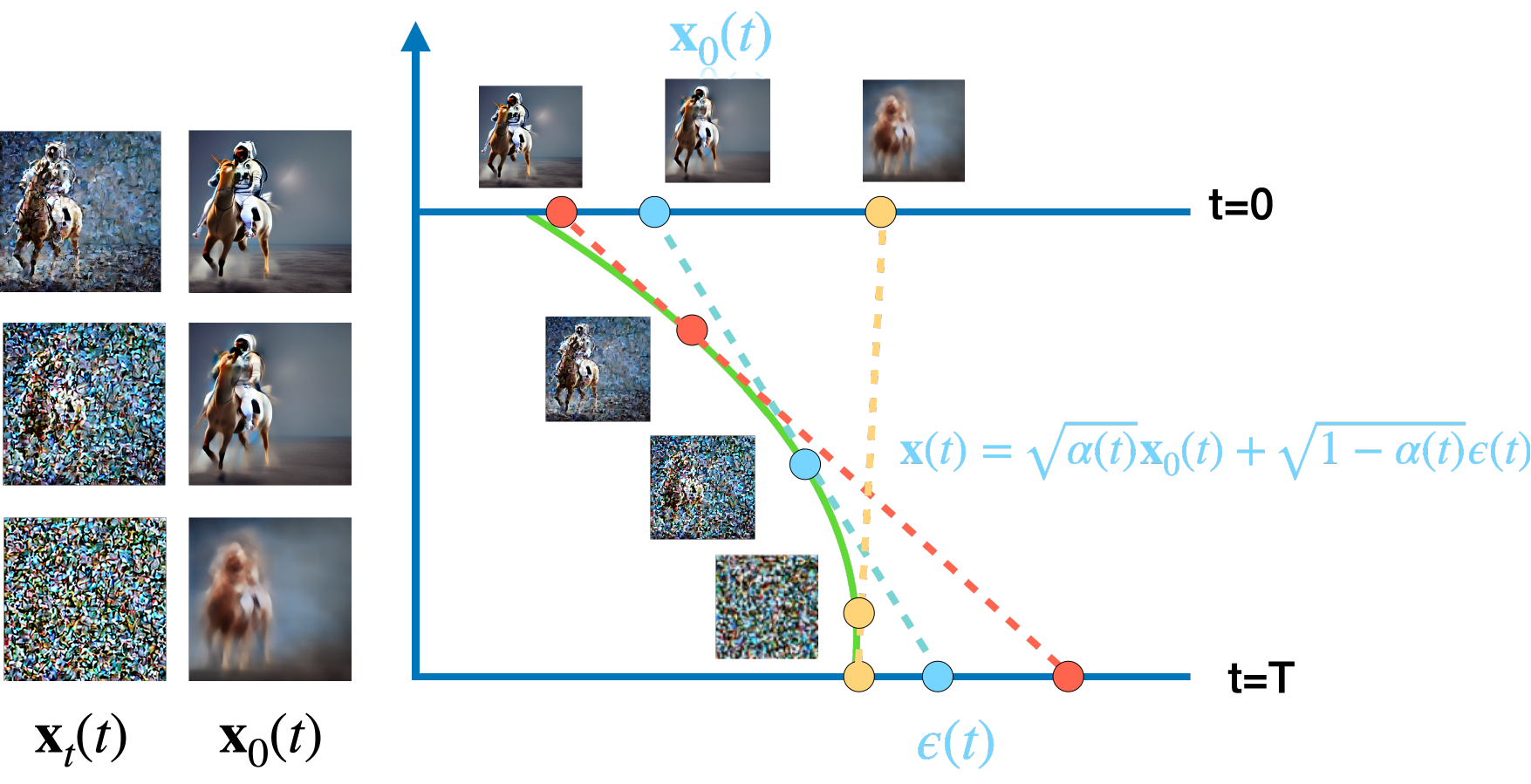}

   \caption{\textbf{DDIM ODE Trajectory.} When noisy image $\mathbf{x}(t)$ is sampled along a DDIM ODE trajectory, there is an induced process in the one-step prediction space $\mathbf{x}_0(t)$.}
   \label{fig: DDIM ODE}
\end{figure}

\begin{figure*}[ht!]
  \centering
    \includegraphics[width=0.9\linewidth]{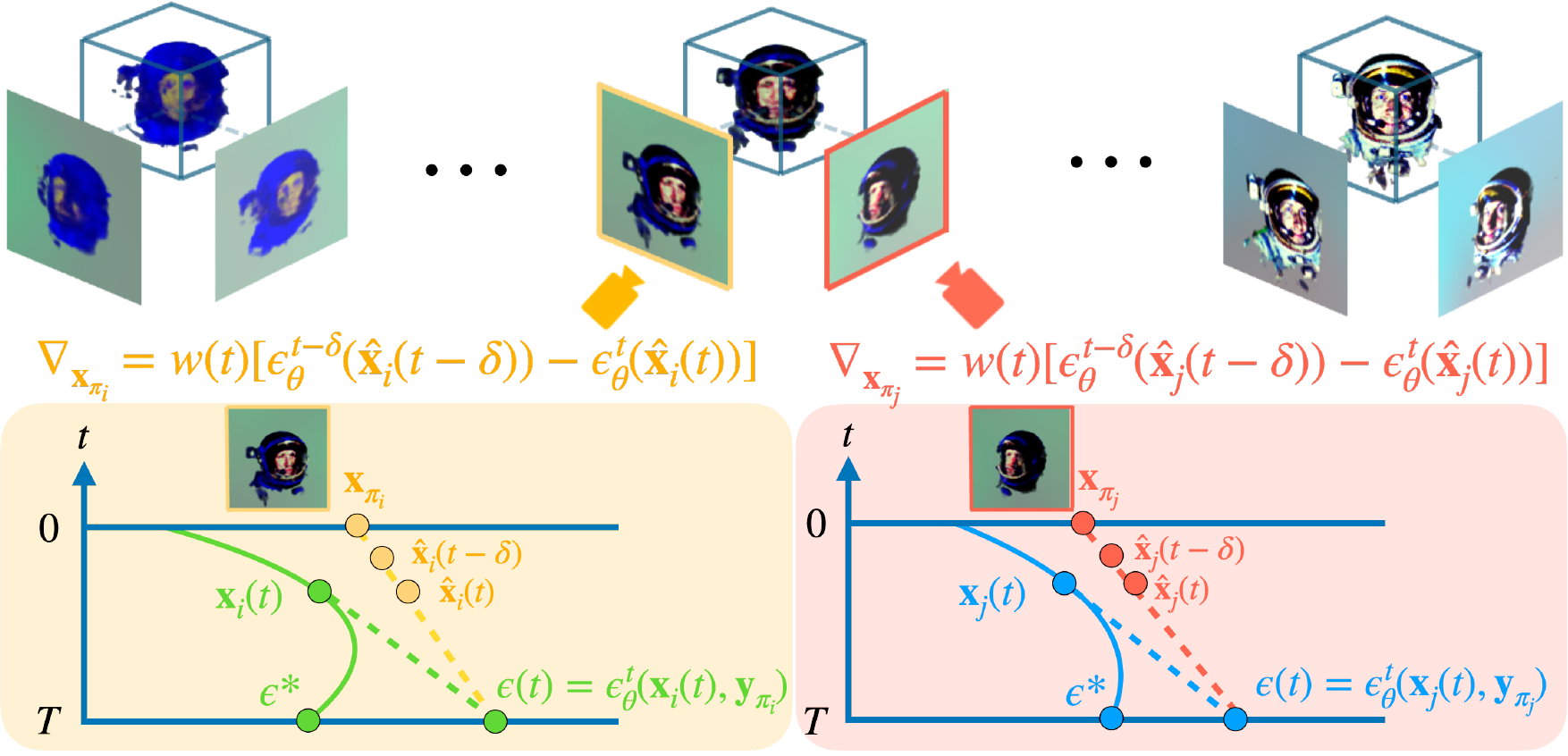}

   \caption{\textbf{Overview of DSD.} A unique ODE starting point $\epsilon^*$ is assigned to each 3D shape throughout the optimization process. Renderings from different views are assumed to be on the view-conditioned ODE, starting from  $\epsilon^*$. At each iteration, DSD simulates the corresponding ODE up to time $t$ and obtains noise prediction $\epsilon(t)$ from the ODE. The rendered view is connected to the ODE by an interpolation approximation, which is then used to obtain the gradient.}
   \label{fig:3D distill method}
\end{figure*}
\section{Related Works}
\label{sec:related}

\vspace{-1mm}
\paragraph{Feed-forward 3D Generation.}

The success of image generative models ~\cite{stablediffusion} has spurred advancements in 3D generation. GAN-based approaches ~\cite{chan2022efficient, gu2021stylenerf} can be trained for 3D generation from images by incorporating rendering inductive bias into the generator. With 3D or multi-view data, diffusion-based 3D or multi-view generators~\cite{ntavelis2023autodecoding, chen2024v3d, shi2023MVDream, kant2024spad} can be trained to produce high-quality results. However, the scale of 3D datasets ~\cite{deitke2023objaverse, reizenstein2021common} remains limited compared to 2D image collections ~\cite{schuhmann2022laion}, which significantly constrains the performance of the trained models.

%-------------------------------------------------------------------------
\vspace{-2mm}
\paragraph{Diffusion-guided Optimization via Score Distillation.}
To circumvent the data limitation, other lines of work seek to optimize a 3D shape by `distilling' prior from a trained 2D diffusion model. The idea is referred to as Score Distillation Sampling (SDS), which is introduced by ~\cite{poole2022dreamfusion, sjc}. However, these methods suffer from over-smoothing and lack of detail. A large Classfier Free Guidance (CFG) ~\cite{CFG} is also needed for effective generation, leading to over-saturation. More importantly, due to the `mode-seeking' behavior of SDS, the 3D shapes are very similar -- an undesirable property for generation tasks. 

Several approaches aim to reduce gradient variance and mitigate over-smoothing, including the use of negative prompts and Classifier-Free Guidance~\cite{NFSD, mcallister2024rethinking, CSD}, prediction differences between adjacent timesteps~\cite{ASD}, time annealing~\cite{HIFA}, and DDIM Inversion~\cite{ISM}. Although these improve fidelity, the lack of diversity in generation persists.
To encourage diversity, ProlificDreamer (VSD)~\cite{VSD} proposes optimizing for a 3D distribution given the prompt by using an additional model to capture the distribution of the current rendering. However, this dual training can be unstable and limits the quality.

Perhaps most closely related to ours, some recent approaches have sought inspiration from ODE-based sampling~\cite{SDI, consistent3d} of diffusion models, but these also do not ensure diversity in their formulation.

\section{Method}

Our goal is to develop a diffusion-guided optimization framework that ensures diversity in the optimized outputs \ie multiple instantiations of the optimization process (using different seeds) should result in different output samples, thus allowing, for example, multiple plausible 3D representations given text conditioning. To develop such a formulation, we take inspiration from the sampling process in denoising diffusion models which inherently yields such diverse samples. We present an initial formulation (\cref{subsec:samplinggradient}) that seeks to enforce the optimization to track the evolution of a diffusion sampling process. In \cref{subsec:dsd}, we then extend this to incorporate scenarios where this might not be exactly feasible (\eg the renderings of a 3D representation being optimized cannot perfectly match target 2D diffusion processes). Before decribing our formulation, we first review some preliminaries about denoising diffusion, score distillation, and DDIM sampling, all of which play a central role in our approach. After outlining our approach, we place our formulation in context of prior score distillation methods, in particular focusing on the diversity of sampling and relation to ODEs (\cref{subsec:framework}).

\subsection{Background}
\label{subsec:prelim}

\textbf{Denoising Diffusion Models (DMs)}. To allow modeling a data distribution $p_0(\mathbf{x})$, diffusion models ~\cite{ddpm, ddim, EDM} are trained to reverse a forward process that gradually adds noise to data to obtain a prior distribution $p_T = \mathcal{N}(0, I)$. In particular, DMs adopt a forward process defined by coefficients $s(t)$ and $\sigma(t)$, where $p(\mathbf{x}(t) | \mathbf{x}(0)) = \mathcal{N}(s(t)~\mathbf{x}(0), (s(t)~\sigma(t))^2~\mathbf{I})$, and train a noise estimator $\epsilon_{\theta}^t(\mathbf{x}_t)$ that can allow sampling from the reverse process \ie mapping noise to data samples.  A common (variance preserving) instantiation of diffusion models that our framework adopts is to set $\sigma(t) = \frac{\sqrt{1-\alpha(t)}}{\sqrt{\alpha(t)}}$ and $s(t)=\sqrt{\alpha(t)}$.

\vspace{1mm}
\noindent \textbf{DDIM Sampling and ODEs.} We build on the DDIM ~\cite{ddim} sampling process that can (deterministically) sample from the distribution modeled by a trained diffusion model $\epsilon_{\theta}$  via the following iterative procedure:
\begin{equation} \label{eqn: ddim sample}
    \frac{\mathbf{x}(t-\delta)}{\sqrt{\alpha(t-\delta)}} = \frac{\mathbf{x}(t)}{\sqrt{\alpha(t)}} + [\sigma(t-\delta) - \sigma(t)] \epsilon_{\theta}^{t}(\mathbf{x}(t), \mathbf{y})
\end{equation}, where $\mathbf{y}$ is a conditioning variable (CLIP text embeddings for the case of text-to-image generation).
This update actually corresponds to a discretized approximation for the  `Probability Flow Ordinary Differential Equation' (PF-ODE) ~\cite{song2020score} for the diffusion model:
\begin{equation} \label{eqn: ddim ode}
    \frac{d\bar{\mathbf{x}}(t)}{dt} = \epsilon_{\theta}^{t}(\sqrt{\alpha(t)}\bar{\mathbf{x}}(t), \mathbf{y})\frac{d\sigma(t)}{dt}
\end{equation} where $\bar{\mathbf{x}}(t) = \frac{\mathbf{x}(t)}{\sqrt{\alpha(t)}}$. In particular, the distribution of generated samples $\mathbf{x}(0)$ obtained by following paths from random starting points $\mathbf{x}(T) \sim \mathcal{N}(0, I)$ is equivalent to the distribution captured by (stochastic) generation from the diffusion model.

\begin{figure*}[!h]
  \centering
    \includegraphics[width=0.8\linewidth]{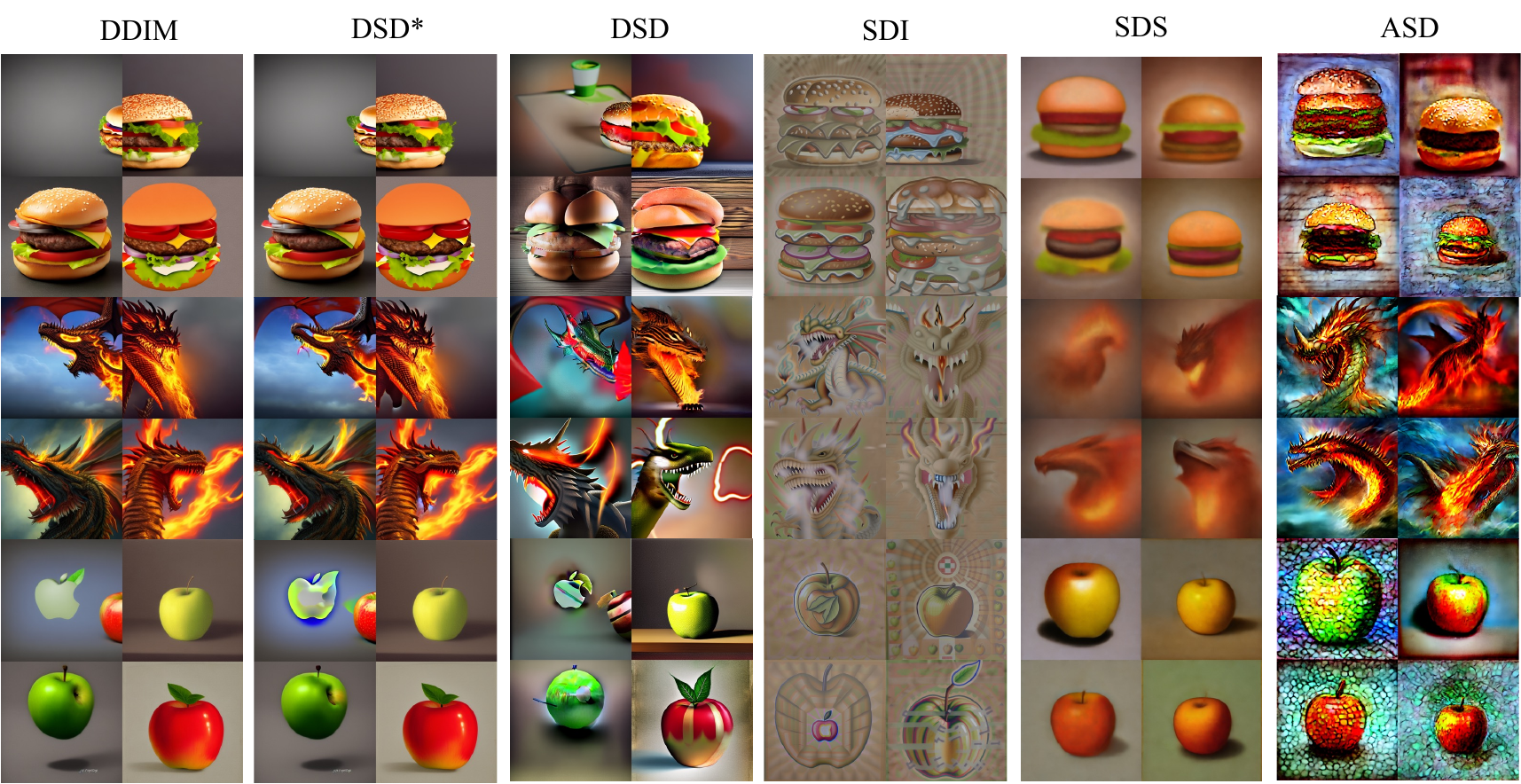}

   \caption{{2D Distillation Results using DDIM as reference.} The prompts are ``A hamburger", ``A dragon with flames coming out of its mouth" and ``An apple". We assign a fixed initial noise for each grid. When the number of DDIM steps equals the optimization step, our method (DSD*) resembles the DDIM sample. When different, the optimized image will deviate from the original ODE by a small margin. We observe that \ours\ yields more diverse and plausible generations compared to alternates.}
   \label{fig:2D distill comp}
\end{figure*}

\noindent\textbf{{Data and Noise Trajectories Induced by DDIM.}} We can define a `one-step data prediction' variable as: 

\begin{equation} \label{eqn: x0defn}
    \mathbf{x}_0(t) = \bar{\mathbf{x}}(t) - \sigma(t)\epsilon(t)
\end{equation}
, where $\epsilon(t) = \epsilon_{\theta}^{t}(\mathbf{x}(t), \mathbf{y})$. Combining \cref{eqn: ddim ode} and \cref{eqn: x0defn}, we can define an induced ODE in $\mathbf{x}_0(t)$ space: 

\begin{equation}\label{eqn: x0 ode}
    \frac{d\mathbf{x}_0(t)}{dt} = -\sigma(t) \frac{d}{dt}\epsilon_{\theta}^{t}(\mathbf{x}(t), \mathbf{y}).    
\end{equation}

A discrete update for this one-step data prediction trajectory can also be derived as:
\begin{equation} \label{eqn: ddim in x0}
    \mathbf{x}_0(t-\delta) = \mathbf{x}_0(t) - \sigma(t-\delta)[\epsilon_{\theta}^{t-\delta}(\mathbf{x}(t-\delta), \mathbf{y}) - \epsilon_{\theta}^{t}(\mathbf{x}(t), \mathbf{y}) ]
\end{equation}
As noticed in ~\cite{SDI, EDM}, the updates in $\mathbf{x}_0$ space have smaller variance compared to that of $\mathbf{x}_t$, which is desirable for an optimization process. 
We visualize the ODE trajectory for $\mathbf{x}(t)$ and the induced ones for $\epsilon(t)$ and $\mathbf{x}_0(t)$ in \cref{fig: DDIM ODE}.

\noindent\textbf{Score Distillation.} To leverage pre-trained diffusion models for guiding optimization, a score distillation-based approach relies on obtaining gradients for an image. We use $\mathbf{x}_{\pi}$ to represent the image whose gradient ($\nabla \mathbf{x}_{\pi}$) will be used to update the optimizable parameter $\psi$. For our 2D experiments, where we directly optimize an image, $x_{\pi} = \psi$ and $\partial{g}/\partial{\psi} = I$. For 3D generation, $\psi$ denotes the parameter of the 3D shape, and $x_{\pi}$ is the rendering of 2D images over camera poses $\pi$,  i.e. $x_{\pi} = g(\psi, \pi)$. Here, $g$ is the differential renderer. Using the the chain rule, one can get:
\begin{equation}
    \nabla_{\psi} = \left (w(t) \nabla_{\mathbf{x}_{\pi}} \frac{\partial g}{\partial \psi} \right)
\end{equation}, where $w(t)$ is a weighting term to normalize gradients at various noise levels for stable optimization. The central design choice in this process is how one can obtain the gradient $\nabla \mathbf{x}_{\pi}$ (as well as how to sample the diffusion timestep $t$ for computing the score).

\subsection{Sampling-based Score Distillation} \label{subsec:samplinggradient} 
For simplicity, we first consider a 2D image generation scenario  \ie using score distillation to  optimize  an image given a text description ($\psi \equiv \mathbf{x}_\pi$). We seek to define a gradient update that allows the optimized variable $\psi = \mathbf{x}_\pi$ to follow a $\mathbf{x}_{0}$ ODE trajectory (\cref{eqn: x0 ode}). Such an update would ensure generation fidelity, as the final image would be likely under the diffusion model (as it corresponds to on ODE output), while also allowing diversity by simply sampling different seeds $\epsilon^* \sim \mathcal{N}(0,\mathbf{I})$ as starting points for the ODE. We can define such a gradient by adapting the update rule for $\mathbf{x}_0(t)$ (\cref{eqn: ddim in x0}):

\begin{equation} \label{eqn: dsd naive gradient}
    \nabla_{\mathbf{x}_\pi}^{\text{Sampling}} := \sigma(t-\delta)[\epsilon_{\theta}^{t-\delta}(\mathbf{x}(t-\delta), \mathbf{y}) - \epsilon_{\theta}^{t}(\mathbf{x}(t), \mathbf{y})]
\end{equation} 

To ensure that $\nabla_{\mathbf{x}_\pi}^{\text{Sampling}}$ allows the update of $\mathbf{x}_{\pi}$ to follow one underlying ODE (specified by a random seed $\epsilon^*$), both $\mathbf{x}(t-\delta)$ and $\mathbf{x}(t)$ should satisfy the DDIM PF-ODE. As shown in the appendix, we guarantee this by defining: 
\begin{align} \label{eqn: dsd naive}
    &\mathbf{x}(t) = \texttt{DDIM-Forward}(\epsilon^{*}, \mathbf{y}, T \rightarrow t) \nonumber  \\
    & \epsilon(t) = \epsilon_{\theta}^{t}(\mathbf{x}(t), \mathbf{y}) \nonumber  \\
    & \mathbf{x}_0(t) = (\mathbf{x}(t) - \sqrt{1- \alpha(t)}\epsilon(t))/\sqrt{\alpha(t)} \\
     & \mathbf{x}(t-\delta) = \sqrt{\alpha(t-\delta)} \mathbf{x}_0(t) + \sqrt{1-\alpha(t-\delta)} \epsilon(t) \nonumber
\end{align}, where $\text{DDIM-Forward}(\epsilon^*, \mathbf{y}, T \rightarrow t)$ is the numerical solver for \cref{eqn: ddim ode} from time $T$ to $t$, starting from $\epsilon^*$.

We refer to the formulation in \cref{eqn: dsd naive gradient} and \cref{eqn: dsd naive} as `Sampling-based Score Distillation' as the resulting optimization simulates the ODE sampling process from $\mathbf{x}(T) = \epsilon^*$. In fact, when the number of optimization steps are chosen to be exactly equal to DDIM sampling steps, outputs from this formulation are equivalent to DDIM samples, and we discuss in the appendix how minor modifications to the step size and $\delta$ can yield similar generations if they are not.

\subsection{Distillation via Interpolation Approximation} \label{subsec:dsd}

While we motivated the formulation in \cref{subsec:samplinggradient} via optimization of a 2D representation, we can use the derived score function to also optimize a 3D representation. As shown by Perp-Neg~\cite{perpneg}, images generated from the same seeds with view-conditioned prompts using the orthogonal update rule can be treated as the set of images from one object. Therefore, we use a \emph{common} seed $\epsilon^*$ across all views of the 3D representation (see appendix for empirical justification) with separate view-conditioned prompt $\mathbf{y}_\pi$~\cite{perpneg} to instantiate the ODE that each view $\mathbf{x}_\pi$ should follow. To obtain different 3D shapes $\psi_i$ and $\psi_j$, we can choose $\epsilon^{*}_{i} \neq \epsilon^{*}_{j}$.

However, naively applying this formulation does not produce impressive (or even valid) 3D generations. The key reason for this is that the 3D representation $\psi$ cannot accurately follow the computed score for each viewpoint $\mathbf{x}_\pi$, thus leading the (3D) optimization process to deviate from the (2D) ODE paths. This is particularly challenging because the gradient formulation in \cref{subsec:samplinggradient} only depends on the specified ODE, and not on the current estimate $\mathbf{x}_{\pi}$, thus providing no `correction' mechanism for the 3D representation to generate valid renderings in case of a drift. 

To address this issue, we observe that both $\mathbf{x}(t)$ and $\mathbf{x}(t-\delta)$ in \cref{eqn: dsd naive} can be expressed as linear combinations of $\mathbf{x}_0(t)$ and $\epsilon(t)$. Our key insight is that we can instead re-define them to be an interpolation of $\mathbf{x}_\pi$ and $\epsilon(t)$:
\begin{align} \label{eqn: dsd interpolation}
     &\mathbf{x}(t) = \texttt{DDIM-Forward}(\epsilon^{*},\mathbf{y}, T \rightarrow t) \nonumber \\
    & \epsilon(t) = \epsilon_{\theta}^{t}(\mathbf{x}(t), \mathbf{y}) \nonumber \\
    &\hat{\mathbf{x}}(t) = \sqrt{\alpha(t)} \mathbf{x}_{\pi} + \sqrt{1 - \alpha(t)} \epsilon(t)  \\
    &\hat{\mathbf{x}}(t-\delta) = \sqrt{\alpha(t-\delta)} \mathbf{x}_{\pi} + \sqrt{1 - \alpha(t)} \epsilon(t) \nonumber
\end{align}
In the ideal case, $\mathbf{x}_{\pi}$ evolves similarly to $\mathbf{x}_0(t)$, but when it does not, this alteration allows a correction mechanism as the gradients are dependent on the current optimization variable. We thus generalize our previous formulation for scenarios where the optimization may drift from the ODE (shown in \cref{fig:3D distill method} ). This leads to our Diverse Score Distillation (DSD) gradient:
\begin{equation} \label{dsd}
    \nabla_{\mathbf{x}_{\pi}}^{DSD} = \sigma(t-\delta)[\epsilon_{\theta}^{t-\delta}(\hat{\mathbf{x}}(t-\delta), \mathbf{y}) - \epsilon_{\theta}^{t}(\hat{\mathbf{x}}(t), \mathbf{y})]
\end{equation}, where we obtain $\hat{\mathbf{x}}(t-\delta)$ and $\hat{\mathbf{x}}(t)$ from \cref{eqn: dsd interpolation}. Using this formulation, we can instantiate our DSD-based 3D optimization procedure in \cref{alg: dsd}. 

\begin{algorithm} []
\caption{Diverse Score Distillation in 3D}
\label{alg: dsd}
\begin{algorithmic}[1]
    \STATE Initialization: 3D parameter $\psi$, trained text-to-image diffusion model $\epsilon_{\theta}^{t}$, prompt $\textbf{y}$, set of cameras around 3D shape $\Pi$, differentiable renderer $g$ 
    \STATE $\epsilon^* \sim \textbf{\textit{N}}(0,I)$
    \FOR{$i = 1$ to N} 
        \STATE $t \leftarrow T(1 - i/N)$
        \STATE $\pi \leftarrow \text{Uniform}(\Pi)$
        \STATE $\mathbf{x}_{\pi} \leftarrow g(\psi, \pi)$
        \STATE Compute $\nabla_{\mathbf{x}_{\pi}}^{\text{DSD}}$ (\cref{dsd})
        
        \STATE $\nabla_{\psi} = w(t) \nabla_{\mathbf{x}_{\pi}}^{\text{DSD}} \frac{\partial g}{\partial \psi}$  
    \ENDFOR
    \RETURN $\psi$
\end{algorithmic}
\end{algorithm}
\subsection{Comparing Score Distillation Formulations} \label{subsec:framework}

Our approach operationalized two key insights: a) if the optimization approach follows the evolution of an ODE, we can expect higher fidelity outputs as these are likely under the diffusion model, and b) the ability to control the generation process (via a random seed) can yield diverse generations. Below, we briefly review alternate score distillation frameworks, in particular highlighting whether they follow an ODE and/or can generate diverse output, and also summarize this in in \cref{tab: formulations}.

\noindent\textbf{Score Distillation Sampling (SDS).}

SDS defines $\nabla_{\mathbf{x}_{\pi}}^{\text{SDS}} = \mathbb{E}_{\epsilon} [\epsilon_{\theta}^{t}(\mathbf{x}(t), \mathbf{y}) - \epsilon]$, where $\mathbf{x}(t)$ is defined by adding random noise to $\mathbf{x}_{\pi}$: $\mathbf{x}(t) = \sqrt{\alpha(t)} \mathbf{x}_{\pi} + \sqrt{1- \alpha(t)} \epsilon$. While this corresponds to the gradient of image log-likelihood under diffusion, the gradient update does not correspond to an ODE. Moreover, the randomly sampled $t$ and $\epsilon$ make this a high-variance estimate. 

\noindent\textbf{Asynchronous Score Distillation(ADS)}~\cite{ASD} also samples a random $\epsilon$ every iteration. Unlike SDS, the gradient is calculated from two adjacent timesteps. The gradient is the same as the DDIM update rule for the sampled ODE at each iteration. However, since the ODE varies with each iteration, this approach also cannot obtain diverse results owing to the high variance gradients.

\noindent\textbf{Score Distillation via Inversion (SDI)}~\cite{SDI} (similarly ~\cite{ISM}) seeks to follow the ODE update rule by inverting the current image (assumed to be $\mathbf{x}_0(t)$) back to $x(t)$ using DDIM Inversion ~\cite{ddim}. The inversion process is a good estimation of the current ODE and a low-variance gradient is applied at every iteration. However, the choice of the ODE cannot be specified and solely depends on the (3D) initialization, leading to lack of diversity in the results.

\noindent\textbf{Consistent3D}~\cite{consistent3d} attempts to follow the consistency ODE \cite{song2023consistency} by setting a constant $\epsilon^{*}$. However, this does not define a valid ODE in the space of denoising diffusion models (including StableDiffusion), necessitating a large classifier-free guidance (CFG) \cite{CFG} scale (50-100) for the method to be effective, which in turn reduces diversity. In contrast, our approach operates with a CFG of 7.5, aligning with the standard 2D sampling process.

\begin{table}
  \centering
  \resizebox{1.0\linewidth}{!}{ % Adjust 0.8 to your preference
  \begin{tabular}{@{}lcccccc@{}}
    \toprule
    & SDS & ASD & SDI & Consistent 3D & VSD & DSD  \\
    \midrule
    \textbf{Diversity} & \ding{55} & \ding{55} & \ding{55} & \ding{51} & \ding{51} & \ding{51} \\
    \textbf{ODE-following} & \ding{55} & \ding{55} & \ding{51} & \ding{55} & \ding{55} & \ding{51} \\
    \bottomrule
  \end{tabular}
  }
  \caption{\textbf{Conceptual Comparisons of Score Distillation Methods.} An optimal distillation method should satisfies both criteria.}
  \label{tab: formulations}
\end{table}

\section{Experiments}
We seek to empirically validate the ability of \ours\ to guide diffusion-based optimization and generate diverse and high-fidelity outputs. We first analyze our formulation in the simple but informative task of image optimization and then examine text-to-3D generation. We further ablate some key decision decisions and also highlight the applicability for distillation-based single-view reconstruction.

\begin{figure*}[h]
  \centering
   % \includegraphics[width=1.0\linewidth]{images/main_exp_v1.2.pdf}
   \includegraphics[width=1.0\linewidth]{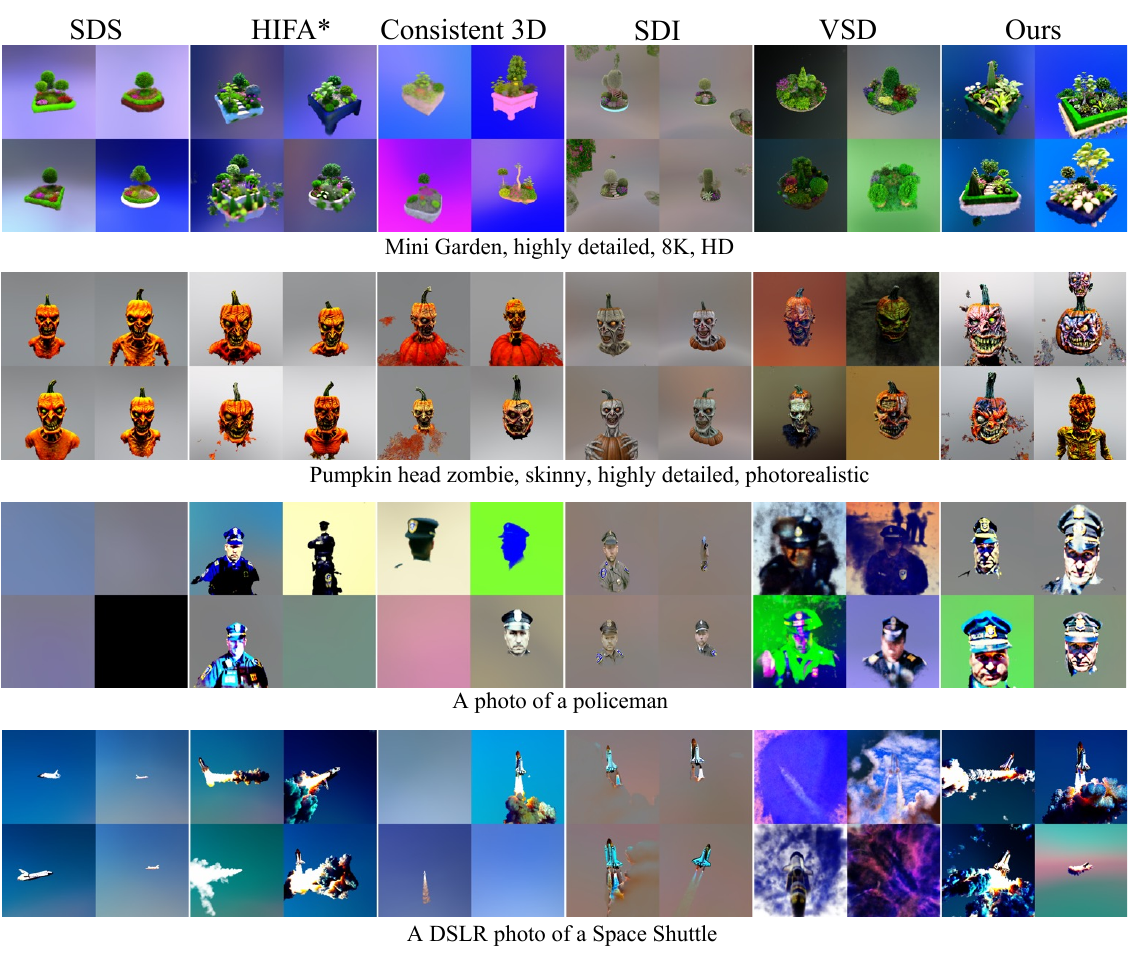}

   \caption{\textbf{Generation Comparison.} We visualize 4 text-to-3D generations from various score distillation methods. We find that DSD is capable of generating high-quality 3D shapes while being more diverse compared to prior methods. Please see supplementary for videos.} %\shubham{Are result figures updated?}}
   \label{fig:3D compare}
\end{figure*}

\paragraph{Baselines.} We select SDS ~\cite{poole2022dreamfusion}, HiFA~\cite{HIFA} and ASD ~\cite{ASD} to represent the methods that sample random noise, and select Consistent 3D ~\cite{consistent3d} and SDI ~\cite{SDI} as the approaches that perform score distillation from an ODE perspective. We also include VSD ~\cite{VSD} as it minimizes the KL divergence between the rendering and image distribution, thus promising diverse generation. For fair comparison across methods, we adopt the codebase from SDI (which builds on a commonly adopted implementation~\cite{threestudio2023}) and use a shared one-stage optimization scheme. We only vary the score function across methods (except for VSD, where we use its released code for multi-particle optimization). 

\subsection{Image Generation via Score Distillation}
In \cref{fig:2D distill comp}, we visualize 4 samples generated by each method across different prompts. When our optimization steps are aligned with the DDIM sampling steps (\ours$^*$), the results are equivalent to DDIM sampling. However, in the general case (\ours), where the optimization steps differ from the DDIM steps (see Appendix), slight variations do arise. Despite this, our method consistently produces generations that are more diverse and of higher fidelity compared to baseline score distillation methods. We use the official code of SDI \cite{SDI} for its 2D results.

\subsection{Text to 3D Generation}
We use various score distillation frameworks to obtain 4 different 3D generations from each text prompt and visualize these in
\cref{fig:3D compare}. 

Consistent with our discussion in \cref{subsec:framework}, SDS \cite{poole2022dreamfusion} and ASD \cite{ASD} suffer from mode-seeking behavior and produce less sharp results. Although VSD can theoretically generate diverse results, simultaneous training of a LoRA~\cite{hu2021lora} network that models the rendering distribution in VSD limits its quality and is potentially unstable. Consistent 3D is more diverse than the aforementioned methods thanks to its ODE formulation. However, the large CFG is harmful to the result. While SDI ~\cite{SDI} is capable of generating high-quality images, its diversity is limited due to the inversion-based ODE seeking. 

% \subsection{Quantitative Results}

We also quantitatively evaluate the quality and diversity of the distilled 3D shapes. For quality, we use FID ~\cite{heusel2017gansfid} and CLIP-SIM ~\cite{CLIP}. For each prompt, FID is calculated across 100 generated samples from Stable Diffusion and the 100 rendered images from 10 3D shapes. For diversity, we propose to use LPIPS ~\cite{lpips} to measure the difference between different 3D shapes generated using the same prompt. We use generated 3D shapes per-prompt and measure the pair-wise LPIPS for the same camera. \cref{tab: main exp} demonstrates the calculated metrics, measured from 15 prompts. 
A smaller FID indicates that the generated 3D shapes are more similar to the 2D images produced by the DM, while a higher CLIP-SIM reflects better text-image alignment of the 3D shapes. Additionally, when the generated 3D shapes appear more distinct, the LPIPS score should be larger. As shown in \cref{tab: main exp}, and consistent with the observations in \cref{fig:3D compare}, our method generates 3D shapes with comparable or superior quality while achieving greater diversity compared to baseline methods.

\begin{table}
  \centering
  \begin{tabular}{@{}lccc@{}}
    \toprule
    & \textbf{FID} $\downarrow$ & \textbf{CLIP-SIM} $\uparrow$ & \textbf{LPIPS} $\uparrow$ \\
    \midrule
    SDS & 270.80 &  27.69 &  0.2695 \\
    HiFA & 258.99 & 28.29 &  0.3202  \\
    % ASD & 261.99 & 27.44 &  0.3356 \\
    SDI & \colorCLIP{70} \textbf{247.98} &  \colorCLIP{40} 28.68 &  0.2407 \\
    Consistent 3D & 270.58 &  28.16 &  0.2936 \\
    VSD & 269.22 &  27.83 &  \colorCLIP{40} 0.3336 \\
    Ours & \colorCLIP{40} 251.40 & \colorCLIP{70} \textbf{28.97} & \colorCLIP{70} \textbf{0.4013} \\
    \bottomrule
  \end{tabular}
  \caption{\textbf{Quantitative Comparison.} We evaluate the fidelity (FID, CLIP-SIM) and diversity (LPIPS) of generations from different methods.}
  \label{tab: main exp}
\end{table}

\paragraph{Ablation.}

\begin{figure}[]
  \centering
   \includegraphics[width=1.0\linewidth]{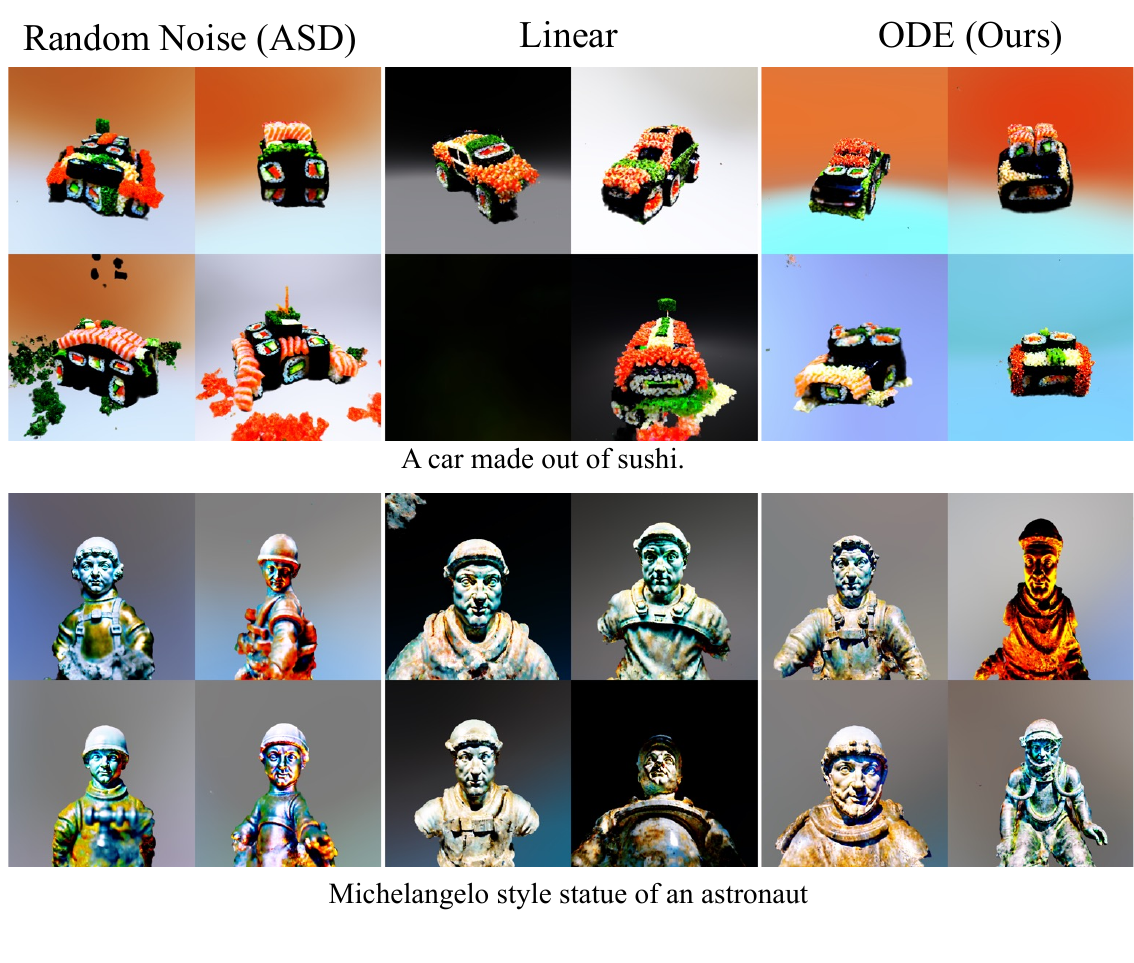}

   \caption{\textbf{Qualitative Ablation Results.} ODE sampling approximation improves diversity and is also more robust compared with Random noise. }
   \label{fig:ablation}
\end{figure}

We also investigate alternative ways of getting the $\mathbf{x}(t)$ and $\mathbf{x}(t-\delta)$ from \cref{eqn: dsd interpolation}:  random noise (similar to ASD~\cite{ASD}), linear approximation (similar to Consistent3D~\cite{consistent3d}, and our sampling with interpolation approximation (\cref{subsec:dsd}). \cref{fig:ablation} and \cref{tab:Ablation} show the qualitative and quantitative results, respectively. It can be observed that simulating a predefined ODE trajectory improves the generation diversity, and that our formulation leads to better results compared to a linear interpolation.

\subsection{Applying DSD to Other Tasks}

\begin{figure}[]
  \centering
   \includegraphics[width=1.0\linewidth]{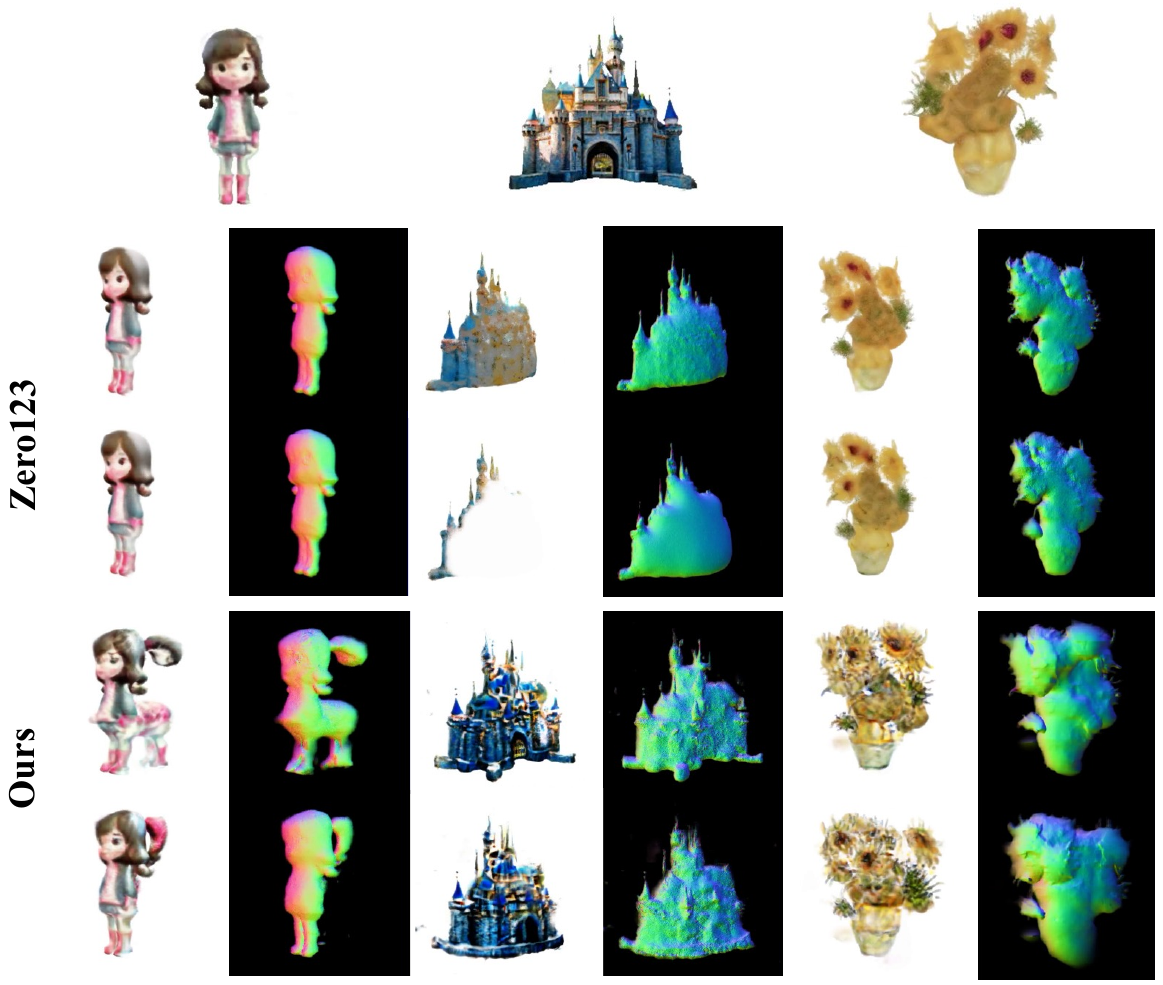}

   \caption{\textbf{Diverse Single-view-to-3D Distillation.} With an image and camera-conditioned DM, DSD reconstructs high-frequency details of the object with diverse interpretations of the underlying geometry. The default distillation adopted by Zero-1-to-3~\cite{zero123} does not yield multi-modal generations and has limited details.}
   \label{fig:single view 3D}
\end{figure}

\noindent{\textbf{Improving Single-view-to-3D Distillation.}}
In addition to text-to-3D generation, score distillation has also been widely used for single-view-to-3D ~\cite{zero123, zhou2023sparsefusion, zhao2024sparseags} reconstruction by leveraging DMs that are trained with image and camera conditions. Most methods still utilize the SDS ~\cite{poole2022dreamfusion} gradient, which is prone to mode-seeking and often results in overly smooth outputs. Given an input image, the geometry of the underlying 3D shape is highly undetermined. 
As shown in \cref{fig:single view 3D}, with the same DM, our method can produce diverse 3D shapes with high-frequency details.

\begin{table}
  \centering
  \begin{tabular}{@{}lccc@{}}
    \toprule
     & \textbf{FID} $\downarrow$ & \textbf{CLIP-SIM} $\uparrow$ & \textbf{LPIPS} $\uparrow$ \\
    \midrule
    {Random (ASD)} & \colorCLIP{40} 265.80 & \colorCLIP{40} 28.24 & 0.3333 \\
    {Linear} &  274.46  &  28.17 & \colorCLIP{40} 0.3579  \\
    {Ours} & \colorCLIP{70} 251.40 & \colorCLIP{70}\textbf{28.97} & \colorCLIP{70} \textbf{0.4013} \\
    \bottomrule
  \end{tabular}
  \caption{\textbf{Ablation.} We ablate three ways of getting $\mathbf{x}(t)$ and $\mathbf{x}(t-\delta)$: random noise at each iteration (ASD \cite{ASD}), linear approximation and ODE sampling approximation. Simulating the given ODE improves the overall quality and diversity. }
  \label{tab:Ablation}
\end{table}

\begin{figure}
    \centering
    \includegraphics[width=1.0\linewidth]{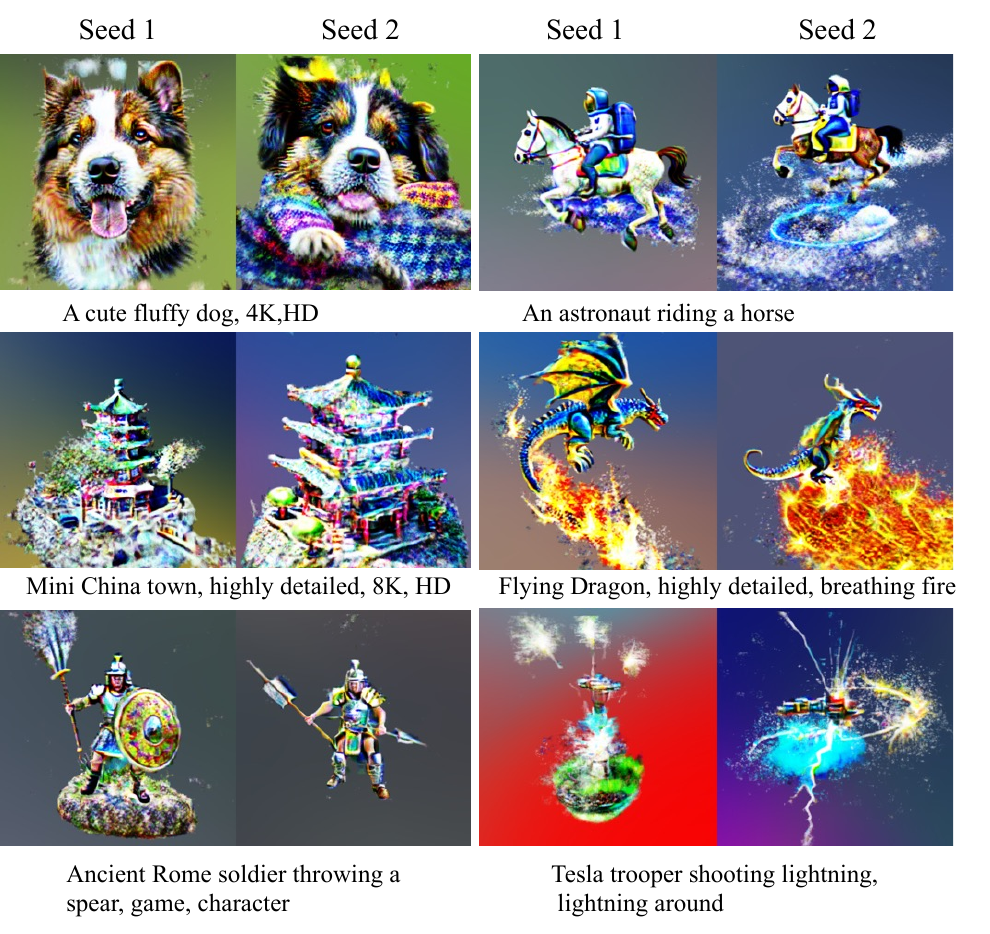}
    \caption{\textbf{Diverse Generation with MVDream.} By using a camera-aware diffusion model, such as MVDream, our method can generate diverse 3D shapes without Janus effects.}
    \label{fig:supp mvdream}
\end{figure}

\noindent\textbf{Diverse Generation with MVDream.} As all other methods ~\cite{ASD, CSD, VSD, SDI, consistent3d, poole2022dreamfusion, HIFA} that use stable diffusion ~\cite{stablediffusion} as the score model, our approach also encounters the Janus effect. One way to mitigate this issue is by employing multiview diffusion models trained on multiview data with camera conditions. For instance, MVDream ~\cite{shi2023MVDream} is such a model, augmented with additional text conditioning. However, similar to DreamFusion ~\cite{poole2022dreamfusion}, the original MVDream suffers from limited diversity due to its reliance on SDS. By integrating our method with MVDream, we achieve diverse 3D generation without the Janus effect. Examples of generated samples are shown in \cref{fig:supp mvdream}.

\section{Discussion}
We presented DSD, a formulation for score distillation that akin to sampling from a diffusion model, can yield diverse samples via optimization, while matching/improving the fidelity compared to existing formulations. However, there are still several challenges and open questions. First, while sampling from diffusion models is (relatively) efficient, score distillation-based optimization is not and it would be interesting to explore formulations that can match the inference speed of denoising diffusion. Moreover, the photo-realism of optimized 3D representations still falls short of generations from the guiding 2D diffusion models, and it remains a challenge to bridge this gap. Finally, we primarily investigated the applications of DSD for text-to-3D and single-view reconstruction, but believe this can be more broadly applicable, for example in 3D editing~\cite{gaussianeditor,instructnerf2nerf} or relighting~\cite{neuralgaffer, litman2025materialfusion}. 
\\
\noindent\textbf{Acknowledgments.} We appreciate the helpful discussions with Lucas Wu, Artem Lukoianov, Ye Hang, Sheng-Yu Wang, Yehonathan Litman and Yufei Ye. This work was supported in part by NSF Award IIS-2345610 and a CISCO gift award.


{
    \small
    \bibliographystyle{ieeenat_fullname}
    \bibliography{main}
}

% \clearpage
% \setcounter{page}{1}
\maketitlesupplementary

\appendix

\section{Additional Discussion}

\textbf{Equivalence of DDIM Sample and Sampling-based Score Distillation}. Here we show the Sampling-based gradient in \cref{eqn: dsd naive} is equivalent to the induced DDIM Sampling process in the single-step prediction $\mathbf{x}_0(t)$ space \cref{eqn: ddim in x0} if the number of optimization steps is equal to DDIM sampling steps. 

We prove this by induction. Assume the equivalence holds true for $t+\delta$, i.e.:
\begin{align} \label{sup_eqn: assumption}
    x(t+\delta) &= \texttt{DDIM-Forward}(\epsilon*, T \rightarrow t+\delta) \nonumber \\
    \epsilon(t+\delta) &= \epsilon_{\theta}^{t+\delta}(x(t+\delta)) \\
     x_0(t+\delta) &= \frac{x(t+\delta) - \sqrt{1-\alpha(t+\delta)}\epsilon(t+\delta)}{\sqrt{\alpha(t+\delta)}} \nonumber
\end{align}
By following update rule for \cref{eqn: dsd naive} with the assumption from \cref{sup_eqn: assumption}, we get:
\begin{equation} \label{subeqn: a}
    x(t) = \sqrt{\alpha(t)} x_0(t+\delta) + \sqrt{1-\alpha(t)}\epsilon(t+\delta) 
\end{equation}
\begin{equation}\label{subeqn: b}
    x_0(t) = x_0(t+\delta) - \alpha(t)[\epsilon_{\theta}^{t}(x(t)) - \epsilon_{\theta}^{t+\delta}(x(t+\delta))]
\end{equation}
From the definition of a single DDIM step, we get 
\begin{align*}
    x^{\texttt{DDIM}}(t) = \sqrt{\alpha(t)} x_0(t+\delta) + \sqrt{1-\alpha(t)}\epsilon(t+\delta). 
\end{align*}
Therefore, we have \begin{align} \label{subeqn: c}
    x(t) &= x^{\texttt{DDIM}}(t) =  \texttt{DDIM-Forward}(\epsilon*, T \rightarrow t)
\end{align}
By plugging \cref{subeqn: a} into \cref{subeqn: b}, we get:
\begin{align} \label{subeqn: d}
     x_0(t) &= \frac{x(t) - \sqrt{1-\alpha(t)}\epsilon_{\theta}^{t}(x(t))}{\sqrt{\alpha(t)}}
\end{align}
As \cref{subeqn: c} and \cref{subeqn: d} resemble our assumption 
at time $t$ (\cref{sup_eqn: assumption}), the induction holds. Here we also include more 2D samples, shown in \cref{supfig: 2D result}. It justifies the proof above empirically (see the first (DDIM) and second (DSD*) column).

\iffalse
Assume: $x_0(t+\delta)$ is the one-step prediction at $t = t+\delta$. \begin{equation}
    x_0(t+\delta) = (x(t+\delta) - \sqrt{1-\alpha(t+\delta)}\epsilon(t+\delta))/\sqrt{\alpha(t)}
\end{equation} \fi

\noindent\textbf{Justification of Using One $\mathbf{\epsilon^*}$ for a 3D Shape.} In 2D, it is trivial that each image corresponds to one ODE starting from $\epsilon^*$. Intuitively, each rendering of a 3D shape should have one unique ODE trajectory, thus more than one $\epsilon$ should be required for one 3D shape. However, finding the set of $\epsilon*$ is highly non-trivial and the set size is infinite. 

\begin{figure}[ht]
    \centering
    \includegraphics[width=0.75\linewidth]{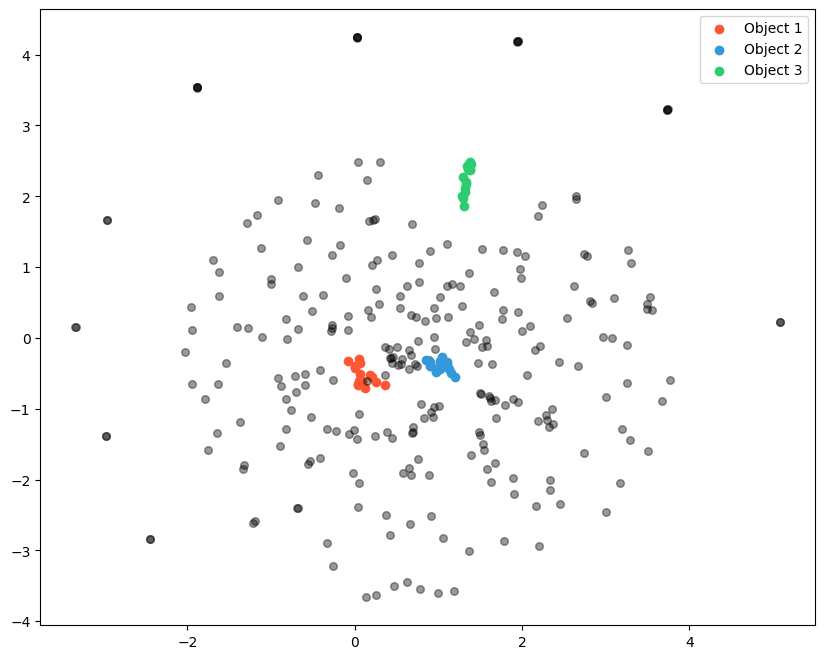}
    \caption{\textbf{DDIM Inverted $\epsilon*$ for different objects, visiluzed using t-SNE.} Here we use DDIM inversion ~\cite{ddim} to obtain the ODE starting points for different objects. Twenty objects are randomly sampled from Objavser ~\cite{deitke2023objaverse} dataset, and we render 16 views for each object. It can be observed that the ODE starting points for the same object are closer to each other. }
    \label{supfig: inversion}
\end{figure}

Empirically however, as shown in \cref{supfig: inversion},  we observe that the $\epsilon^*$ for each object is actually close to each other. This enables us to approximate the set of $\epsilon$ of each object with one ODE starting point. Additionally, our interpolation approximation \cref{eqn: dsd naive} allows slight drift from the ODE, which makes the optimization of 3D feasible. To further reduce the approximation error, we utilize view-conditioned text prompts and the sampling strategy from prep-neg ~\cite{perpneg}. Specifically, for different views, the ODEs will start from the same point but are conditioned with corresponding views.

\begin{algorithm} [!h]
\caption{Diverse Score Distillation in 2D}
\label{alg: detail 2D}
\begin{algorithmic}[1]
    \STATE Initialization: 2D image $\mathbf{x_{\pi}}$, trained text-to-image diffusion model $\epsilon_{\theta}^{t}$, prompt $\textbf{y}$, ddim sampling steps $N_{ddim}$, optimization steps $N$, $\delta = T/N_{ddim}$, lr = $N_{ddim}/N$
    \STATE $\epsilon^* \sim \textbf{\textit{N}}(0,I)$ 
    \FOR{$i = 1$ to N} 
        \STATE $t \leftarrow T(1 - i/N)$
        \STATE $t_{\delta} \leftarrow \min(t + \delta, T)$
        \STATE \textcolor{blue}{$t_{ddim} \leftarrow T(1 - \frac{\ceil{t_{\delta}/N}}{N_{ddim}})$}
        \STATE \textcolor{blue}{$\mathbf{x}(t_{ddim}) \leftarrow \texttt{DDIM-Forward}(\epsilon^{*}, \mathbf{y}, T \rightarrow t_{ddim})$}
        \STATE \textcolor{blue}{$\epsilon(t_{ddim}) = \epsilon_{\theta}^{t_{ddim}}(\mathbf{x}(t_{ddim}))$}
        \STATE $\mathbf{x}(t) = \sqrt{\alpha(t)} \mathbf{x}_{\pi} + \sqrt{1 - \alpha(t)} \textcolor{blue}{\epsilon(t_{ddim})}$
        \STATE $\mathbf{x}(t_{\delta}) = \sqrt{\alpha(t_{\delta})} \mathbf{x}_{\pi} + \sqrt{1 - \alpha(t_{\delta})} \textcolor{blue}{\epsilon(t_{ddim})}$
        \STATE $\nabla_{\mathbf{x}_{\pi}} \leftarrow lr * \sigma(t)[\epsilon_{\theta}^{t}(\mathbf{x}(t), \mathbf{y}) - \epsilon_{\theta}^{t_{\delta}}(\mathbf{x}(t_{\delta}), \mathbf{y})]$
    \ENDFOR
    \RETURN $\psi$
\end{algorithmic}
\end{algorithm}

\begin{figure*}[h]
  \noindent
  \begin{minipage}[t]{0.48\textwidth}
    
\begin{algorithm} [H]
    \caption{DSD}
    \label{alg: DSD supp}
    \begin{algorithmic}[1]
        \STATE Initialization: 3D parameter $\psi$, trained text-to-image diffusion model $\epsilon_{\theta}^{t}$, prompt $\textbf{y}$, set of cameras around 3D shape $C$, differential renderer g 
        \STATE \textcolor{blue}{$\epsilon^* \sim \textbf{\textit{N}}(0,I)$}
        \FOR{$i = 1$ to N} 
            \STATE \textcolor{blue}{$t \leftarrow T(1 - i/N)$}
            \STATE $\pi \leftarrow \text{Uniform}(\Pi)$
            \STATE $\mathbf{x}_{\pi} \leftarrow g(\psi, \pi)$
            \STATE $\mathbf{x}(t+\delta) = \texttt{DDIM-Forward}(\epsilon^{*},\mathbf{y}, T \rightarrow t+\delta) \nonumber$
            \STATE \textcolor{blue}{$\epsilon(t+\delta) = \epsilon_{\theta}^{t+\delta}(\mathbf{x}(t+\delta), \mathbf{y})$}
            \STATE \textcolor{blue}{$\hat{\mathbf{x}}(t+\delta) = \sqrt{\alpha(t+\delta)} \mathbf{x}_{\pi} + \sqrt{1 - \alpha(t+\delta)} \epsilon(t+\delta)$}
            \STATE \textcolor{blue}{$\hat{\mathbf{x}}(t) = \sqrt{\alpha(t)} \mathbf{x}_{\pi} + \sqrt{1 - \alpha(t)} \epsilon(t+\delta)$}
           
            \STATE $\nabla_{\psi} = w(t) [\epsilon_{\theta}^{t}(\hat{\mathbf{x}}(t), \mathbf{y}) - \epsilon_{\theta}^{t+ \delta}(\hat{\mathbf{x}}(t + \delta), \mathbf{y})]\frac{\partial g}{\partial \psi}$  
        \ENDFOR
        \RETURN $\psi$
    \end{algorithmic}
    \end{algorithm}

  \end{minipage}
  \hfill
  \begin{minipage}[t]{0.48\textwidth}

\begin{algorithm} [H]
    \caption{SDS}
    \label{alg: sds}
    \begin{algorithmic}[1]
        \STATE Initialization: 3D parameter $\psi$, trained text-to-image diffusion model $\epsilon_{\theta}^{t}$, prompt $\textbf{y}$, set of cameras around 3D shape $C$, differential renderer g 
        \STATE \textcolor{red}{No fixed sample of $\epsilon$}
        \FOR{$i = 1$ to N} 
            \STATE \textcolor{red}{$t \leftarrow \text{Uniform}(1, T)$}
            \STATE $\pi \leftarrow \text{Uniform}(\Pi)$
            \STATE $\mathbf{x}_{\pi} \leftarrow g(\psi, \pi)$
            \STATE \textcolor{red}{$\epsilon \sim \textbf{\textit{N}}(0,I)$}
            \STATE \textcolor{red}{$\mathbf{x}(t) \leftarrow  \sqrt{\alpha(t)} \mathbf{x}_{\pi} + \sqrt{1 - \alpha(t)} \epsilon $}
            
            \STATE $\nabla_{\psi} = w(t) [\epsilon_{\theta}^{t}(\mathbf{x}(t), \mathbf{y}) - \epsilon] \frac{\partial g}{\partial \psi}$  
        \ENDFOR
        \RETURN $\psi$
    \end{algorithmic}
    \end{algorithm}

  \end{minipage}
\end{figure*}

\section{Implementation Details}

\textbf{Diverse Score Distillation in 2D.} In 2D, when the number of DDIM sampling steps is not equal to the optimization steps, we can make slight modifications to \cref{eqn: dsd naive} and yield similar generation results (see the third column in \cref{supfig: 2D result}). The algorithm is shown in \cref{alg: detail 2D}. The core idea is to simulate the discrete DDIM solver as closely as possible using the fixed DDIM sample timesteps $t_{ddim}$. For example, we will use the set $\{1000, 900, \dots, 100\}$ when $N_{ddim} = 10$. 

\noindent\textbf{Diverse Score Distillation in 3D.} \cref{alg: DSD supp} shows our method for 3D generation side by side with SDS ( \cref{alg: sds}). We implement our code on Threestudio~\cite{threestudio2023}, a framework for score distillation. We use a maximum of 10 DDIM sampling steps during the optimization process. Our choice of $\delta$ is similar to ~\cite{ASD}, where $\delta = 0.1(t-t_{\texttt{MIN}})$. We use a cfg of 7.5 for $\epsilon^t_{\theta}(\mathbf{x}(t), y)$ and 1.0 for $\epsilon^{t+\delta}_{\theta}(\mathbf{x}(t+\delta), y)$, as we find this provides a better guidance in 3D. During the ODE-solving process, we use a constant cfg of 7.5. Our choice of 3D representation and geometrical regularization is identical to that of SDI ~\cite{SDI} across experiments. 

\section{More 2D Results}

\begin{figure*}
    \centering
    \includegraphics[width=0.75\linewidth]{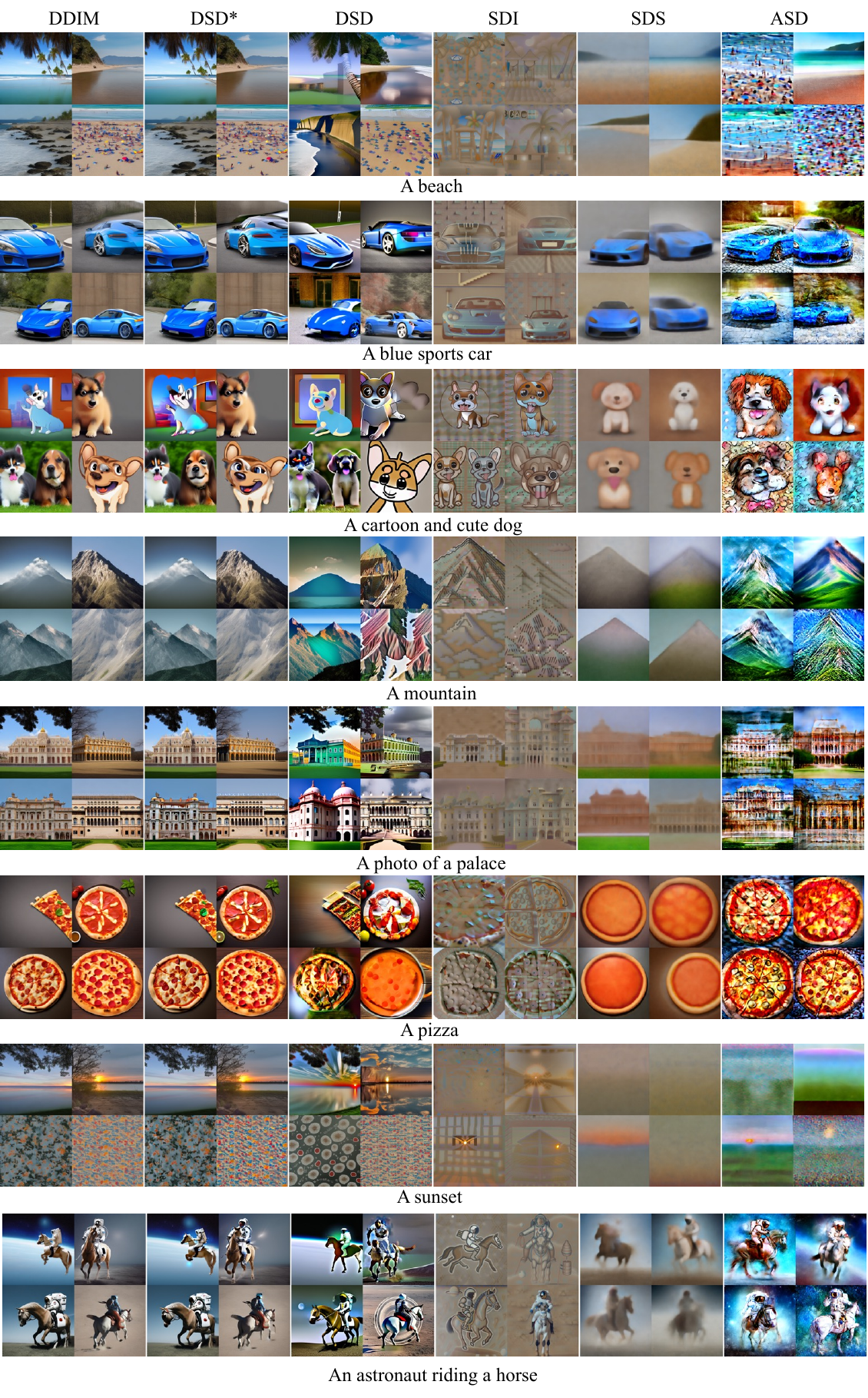}
    \caption{\textbf{More 2D Distillation Results.} Our method can simulate the original ODE perfectly (DSD*) when the number of DDIM steps and optimization steps are equal. Ours (DSD) also generates diverse results by stimulating the underlying ODEs, while other baselines ~\cite{SDI, poole2022dreamfusion, ASD} have no explicit guarantee of diverse generation.}
    \label{supfig: 2D result}
\end{figure*}

\noindent\textbf{More 2D Results.} Additional 2D distillation results are presented in \cref{supfig: 2D result}. Our method achieves greater diversity in the generated outputs by simulating the underlying ODE, whereas the baselines provide no explicit guarantee of diverse generation.

\begin{figure}
    \centering
    \includegraphics[width=0.9\linewidth]{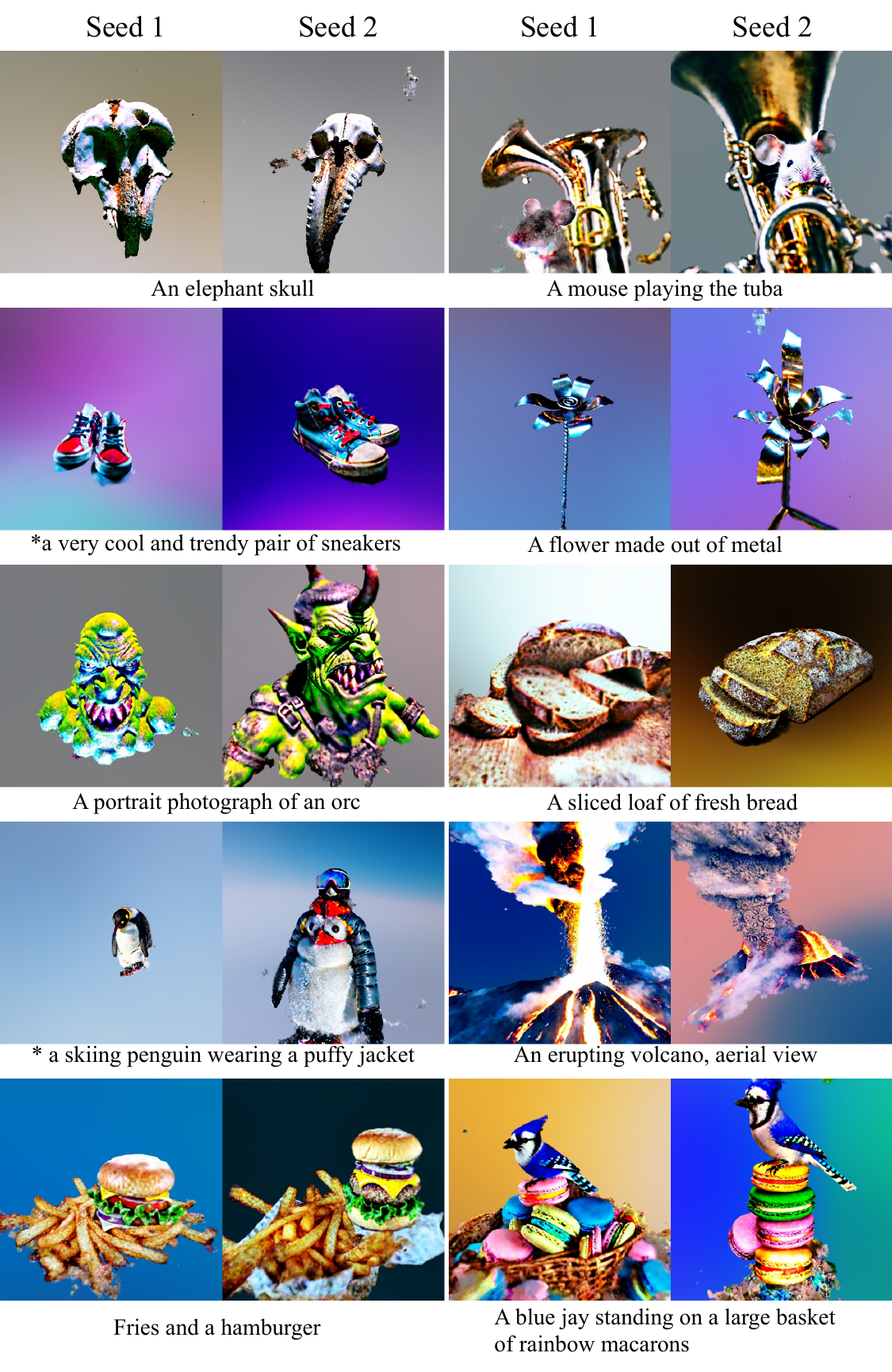}
    \caption{\textbf{More 3D generation Results.} As different 3D shapes are generated with different ODEs, the results are diverse.  * "A DSLR photo of". }
    \label{supfig: 3D result}
    \vspace{-1cm}
\end{figure}

\noindent\textbf{More 3D Results.} We include more diverse 3D generation results in \cref{supfig: 3D result}. Our method is capable of generating high-quality 3D shapes while maintaining diversity.

%\clearpage
%\input{sec/3_method}

\end{document}